%% file: main.tex
\documentclass[letterpaper, 10 pt, conference]{ieeeconf}  

\IEEEoverridecommandlockouts                              

\usepackage{cite}
\usepackage{amsmath,amssymb,amsfonts}
\usepackage{algorithm,algpseudocode}
\usepackage{caption}
\usepackage{subcaption}
\usepackage{multirow}
\usepackage{graphicx}
\usepackage{textcomp}
\usepackage{url}
\usepackage{xcolor}
\usepackage{xspace}
\usepackage{comment}
\usepackage{hyperref}

\newcommand{\etal}{et al.~}

\newcommand{\armor}{{\texttt{ARMOR}}\xspace}

\title{\LARGE \bf
\armor: Attack-Resilient Reinforcement Learning Control for UAVs
}

\author{
Pritam Dash$^{\dagger}$, Ethan Chan$^{\dagger}$, Nathan P. Lawrence$^{*}$, Karthik Pattabiraman$^{\dagger}$\\
$^{\dagger}$University of British Columbia, Canada.  
$^{*}$University of California, Berkeley, USA\\
$^{\dagger}$\{pdash, echan, karthikp\}@ece.ubc.ca.
$^{*}$nplawrence@berkeley.edu
}

\begin{document}

\maketitle
\thispagestyle{empty}
\pagestyle{empty}

\begin{abstract}
Unmanned Aerial Vehicles (UAVs) depend on onboard sensors for perception, navigation, and control. However, these sensors are susceptible to physical attacks, such as GPS spoofing, that can corrupt state estimates and lead to unsafe behavior. While reinforcement learning (RL) offers adaptive control capabilities, existing safe RL methods are ineffective against such attacks. 
We present \armor (\underline{A}daptive \underline{R}obust \underline{M}anipulation-\underline{O}ptimized State \underline{R}epresentations), an attack-resilient, model-free RL controller that enables robust UAV operation under adversarial sensor manipulation. 
Instead of relying on raw sensor observations, \armor learns a robust latent representation of the UAV’s physical state via a two-stage training framework. In the first stage, a teacher encoder, trained with privileged attack information, generates attack-aware latent states for RL policy training. 
In the second stage, a student encoder is trained via supervised learning to approximate the teacher’s latent states using only historical sensor data, enabling real-world deployment without privileged information.
Our experiments show that \armor outperforms conventional methods for ensuring UAV safety. 
Further, \armor improves generalization to unseen attacks and reduces training cost by eliminating the need for iterative adversarial training. 

\end{abstract}

\section{Introduction}
\label{sec:intro}
\input{sections/1-introduction}
\section{Related Work}
\label{sec:background}
\input{sections/2-background}

\section{Preliminaries}
\label{sec:prelims}
\input{sections/3-threat-model}
\section{\armor: Two-Staged Training Framework}
\label{sec:design}

\input{sections/4-design}

\section{Evaluation and Results}
\label{sec:results}

\input{sections/5-results}
\section{Discussions}
\label{sec:discussions}
\input{sections/6-discussions}

\section{Conclusions}
\label{sec:conclusions}
\input{sections/7-conclusions}

\section*{Acknowledgements}
\begin{footnotesize}
    This work was partially supported by the Natural Sciences and Engineering Research Council of Canada (NSERC), the Department of National Defence (DND) Canada, and a Four Year Fellowship (4YF) from UBC. 
\end{footnotesize}

\bibliographystyle{IEEEtran}
\bibliography{bibliography}

\end{document}

%% file: sections/1-introduction.tex
\begin{figure}
    \centering
    \includegraphics[width=\linewidth]{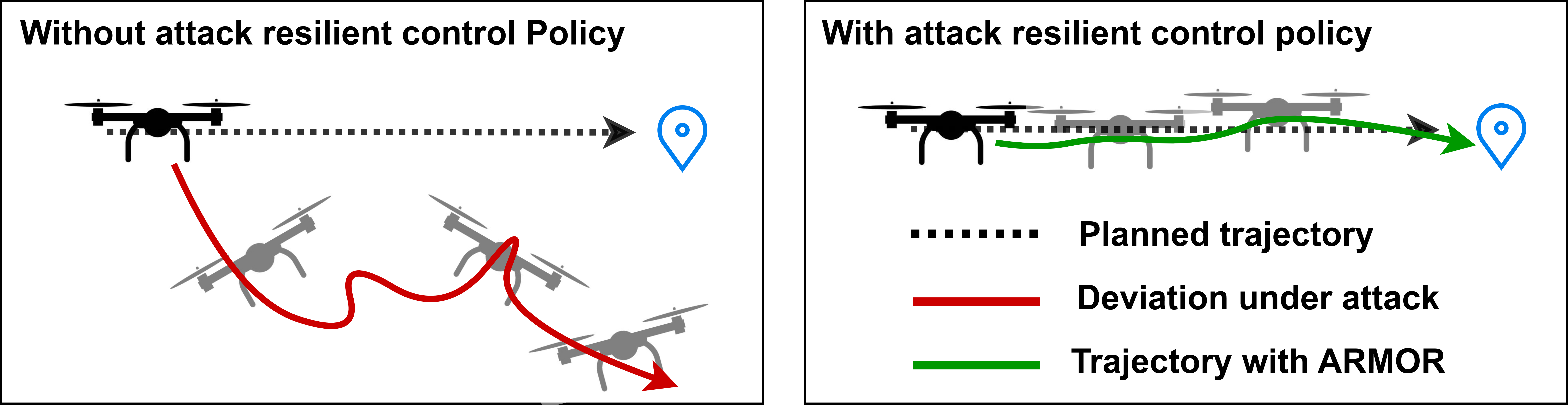}
    \caption{\textbf{Left:} Without an attack-resilient control policy, a UAV subjected to physical attacks deviates significantly from its planned trajectory, leading to mission failure. \textbf{Right:} With an attack-resilient control policy like \armor, the UAV remains on course despite attacks and completes the mission.}
    \label{fig:motivation}
\end{figure}

Unmanned Aerial Vehicles (UAVs) are extensively used in various applications including logistics, agriculture, surveillance, and emergency services~\cite{rav-industry}. 
UAVs rely on onboard sensors for perception, autonomous navigation, and control. 
Correctness of sensor measurements is critical to achieving safe and reliable performance in UAV missions.
However, sensors are susceptible to {\em physical attacks} launched by injecting malicious signals or noise through the physical channel. 
Examples of such attacks are GPS spoofing~\cite{gpsspoofing1}, gyroscope tampering using acoustic noise~\cite{gyroscopespoofing}, and optical sensor spoofing through laser beams~\cite{opticalspoofing}. 
Physical attacks corrupt a UAV's physical state estimates, leading to unsafe control actions and resulting in deviations from planned trajectories, or crashes, as illustrated in Figure~\ref{fig:motivation}.

Model-free Reinforcement Learning (RL) has emerged as a promising approach for UAV control, enabling adaptive decision-making in complex and dynamic environments~\cite{deeprl-control}. 
However, as RL-based controllers also rely on sensors, they are vulnerable to physical attacks.
Techniques like shielding~\cite{rl-sheilding} and control barrier functions (CBF)~\cite{rl-cbf} have been proposed for safe policy learning. 
However, they are not effective under physical attacks. 
Shielding and CBFs rely on prior definitions of unsafe actions and well-defined boundaries of the unsafe action space.
Physical attacks present a fundamentally different threat model. 
They can cause the controller to execute unsafe actions under the illusion that they remain within safe limits. 
For example, GPS spoofing can cause incremental deviations during a UAV mission -  
these deviations might appear safe within the defined action space, but they can cumulatively cause the UAV to follow an unintended and potentially dangerous path~\cite{gpsspoofing2, pid-piper, specguard}. 

Adversarial training is a popular approach for developing attack-resilient RL-based controllers~\cite{rarl, adv-train-deeprl}. 
However, adversarial training has three limitations under physical attacks~\cite{limit-adv-train-deeprl, specguard}. 
(1) It incurs a high training cost due to the iterative generation of adversarial scenarios. (2) Lacks generalizability, as the policy is only effective against the specific attack patterns encountered in training. 
(3) It lacks zero-shot effectiveness against previously unseen attacks.

We propose \armor, an attack-resilient RL-based controller for UAVs. Instead of directly relying on high-dimensional physical state information from onboard sensors, \armor generates a robust latent representation of the UAV's physical state that is  designed to withstand physical attacks. 
This latent state representation allows the UAV to operate safely and complete its missions, despite the malicious interventions.
Our main innovation is a two-stage offline training framework. In the first stage, we train the RL policy using a {\em teacher encoder} that uses privileged information. In the second stage, we adapt the RL policy using a {\em student encoder}, which relies solely on the onboard sensors. 
During online deployment, only the student encoder is used. 

The teacher encoder has access to privileged information, such as the target sensor under attack, the magnitude of the sensor bias, and the duration of sensor manipulation. 
By combining the UAV's high-dimensional physical state information with the privileged information, the teacher encoder generates a robust latent state representation using a Variational Autoencoder (VAE)~\cite{vae}. 
This latent representation allows the RL policy to achieve high performance in control tasks, and it also remains resilient to physical attacks. 

Since privileged information is unavailable in real-world deployment scenarios, we introduce a student encoder that relies solely on the onboard sensors.
The student encoder processes the UAV's historic physical state information derived from the onboard sensors using a Long Short Term memory (LSTM) network, capturing temporal dependencies to generate a robust latent state representation. 
The student encoder learns to approximate the latent state representation of the teacher encoder through supervised learning. 
The RL-policy is used with the student encoder for online inference. 

By decoupling the learning process into teacher and student encoders and leveraging privileged information, \armor eliminates the need for iterative adversarial scenario generation, resulting in significant reductions in training costs. 
The robust latent state representation further enhances \armor's generalizability across a wide range of attack types. 
In addition, \armor demonstrates zero-shot generalization, where a policy trained on one attack type remains robust to previously unseen attack modalities without retraining.

While prior work has explored the use of privileged information~\cite{priviledged-info} in robotics, it focuses on enhancing control under normal conditions, rather than under physical attacks. 
In contrast, our work designs robust latent state representations specifically optimized for control under adversarial perturbations. 
We have released \armor's code at \url{https://github.com/DependableSystemsLab/armor}.



\textbf{\em{Contributions.}}
Our contributions are as follows: 
\begin{itemize}
    \item We introduce a two-stage offline training framework for developing an attack-resilient model-free RL controller for UAVs. First, the controller is trained using privileged information to enable robust and efficient policy learning. Second, the policy is adapted for online deployment using only onboard sensor data. 
    \item We propose a robust state representation method that transforms the UAV's high-dimensional physical state information into a resilient latent vector representation, ensuring robustness to physical attacks.
    \item We propose a transfer learning strategy that enables the RL controller to infer robust latent state representations without using privileged information, and instead relying solely on historical sensor information. 
\end{itemize}


%% file: sections/2-background.tex
\subsection{Safe and Resilient Model-free Policy Learning}
\label{sec:rw-safe-policy}

Prior work in safe RL focuses on uncertainty, typically by constraining actions to remain within a safe set~\cite{rl-saferl, res-nav-uncertainity, control-uncertain,  safe-robot-manipulation}. 
A few prominent examples are shielding~\cite{rl-sheilding}, control barrier functions (CBF)\cite{rl-cbf}, and safety critics~\cite{safety-critic}. 
However, physical attacks manipulate the UAV's perception of its state, leading to unsafe actions that appear safe within the defined action space. 
The above safe RL mechanisms are designed to handle unsafe actions under normal operating conditions, and they are not designed to mitigate deliberate state manipulations caused by physical attacks.

Robust RL techniques are another category of work proposed to handle state manipulations.  
However, these methods attempt to handle disturbances by learning conservative policies~\cite{robust-rl} or training against adversarial perturbations~\cite{rarl, adv-train-deeprl}. 
While they improve robustness, they rely on predefined uncertainty sets or known attack patterns, limiting generalization to unseen attacks. 
Moreover, adversarial training increases training cost, and broad uncertainty sets often produce overly conservative policies that degrade performance.


\subsection{UAV State Representation}
\label{sec:rw-state-rep}

Prior work has explored contrastive learning for robot state representation~\cite{con-learn-robots}, which learns discriminative representations, typically supervised, but has seen limited use in safety-critical robotics. 
Lee et al.\cite{priviledged-info} introduced latent vectors with privileged information for quadrupeds, though under nominal sensor conditions rather than attacks. 
Our work builds on these foundations but focuses on resilient control under adversarial conditions, addressing a critical gap.

%% file: sections/3-threat-model.tex
UAVs rely on sensors for perception. For instance, the GPS measures position ($x, y, z$), the gyroscope measures angular orientation ($\phi, \theta, \psi$), the accelerometer measures velocity ($\dot{x}, \dot{y}, \dot{z}$) and acceleration ($\ddot{x}, \ddot{y}, \ddot{z}$), the magnetometer measures heading direction, the barometer measures altitude ($z$), and the optical flow sensor measures horizontal motion. 

\begin{figure*}[t]
    \centering
    \includegraphics[width=0.7\linewidth]{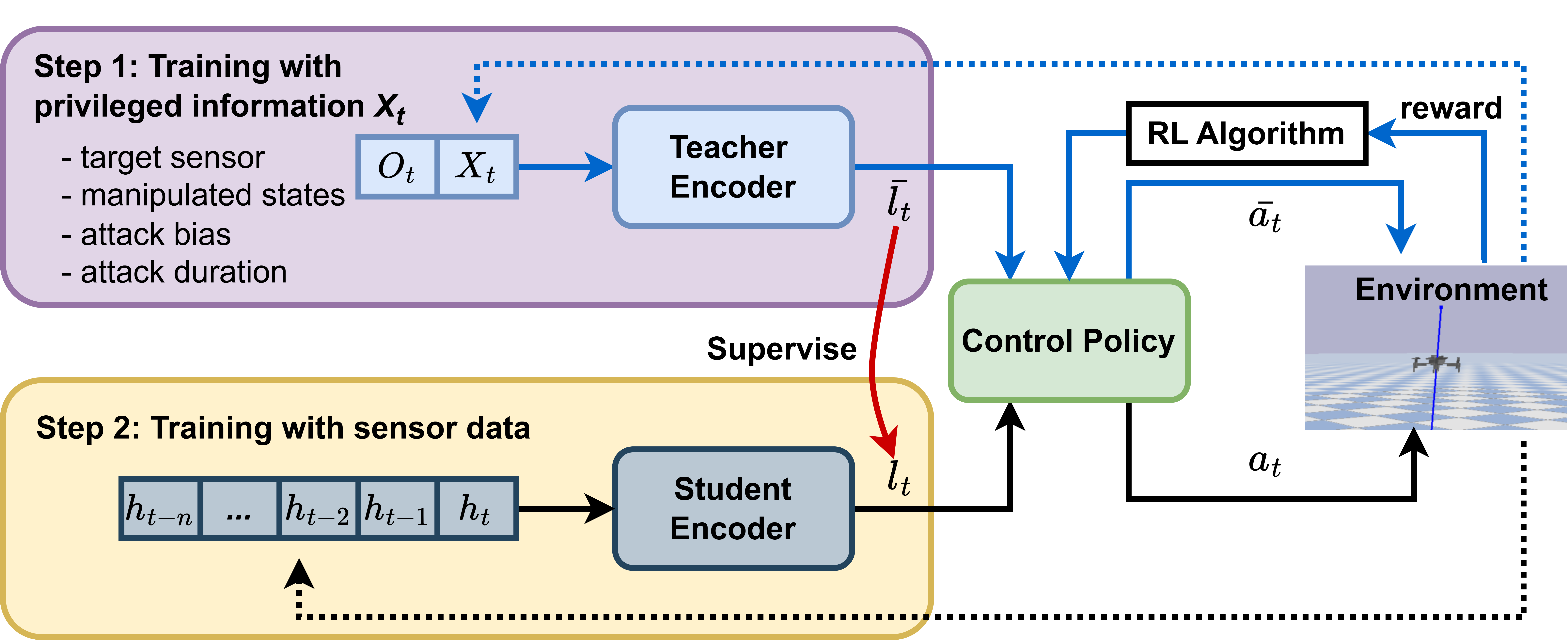}
    \caption{\textbf{Overview of \armor's two-staged training approach}. First, a {\em teacher encoder} is trained with privileged information that includes attack information---target sensor, corrupted states, attack duration, etc. The {\em control policy} is trained jointly with the teacher encoder. 
    Second, a {\em student encoder} is trained to approximate the teacher encoder via supervised learning. The student encoder does not have access to privileged information, instead, it relies on a stream of historic physical state information derived from onboard sensors. 
    For online deployment, the control policy uses the student encoder.}
    \label{fig:design}
\end{figure*}

\subsection{Threat Model}
\label{sec:threat-model}

Physical attacks introduce bias into sensors that propagates through the UAV's feedback control loop, corrupting the UAV's physical state estimates, and subsequently results in erroneous actuator signals~\cite{ravage}, leading to unsafe consequences such as collisions or crashes~\cite{pid-piper}.

We assume an adversary capable of launching GPS spoofing~\cite{gpsspoofing2}, gyroscope and accelerometer tampering via acoustic interference~\cite{injected-delivered}, magnetometer corruption with electromagnetic signals~\cite{emi-attack-drone}, and optical flow spoofing~\cite{opticalspoofing}. 
We assume the adversary can inject attacks of varying bias, intensity, and patterns.
The adversary can also launch stealthy attacks to disrupt the UAV~\cite{stealthy-attacks} gradually. These attacks have been shown to be practical in real-world settings~\cite{physical-attacks}. 
%
We assume the adversary operates from fixed locations and launches attacks intermittently during the UAV mission. However, they cannot compromise actuators, or the onboard software. This threat model is in line with other work in the area~\cite{specguard, pid-piper, physical-attacks}.

We define zero-shot generalization as the ability of the RL control policy trained under one attack modality (e.g., GPS spoofing) to remain effective against previously unseen modalities (e.g., gyroscope or accelerometer tampering) without retraining or fine-tuning. 


\subsection{UAV Control Design}
\label{sec:control-design}


The control architecture for a UAV consists of two primary components: motion generation and tracking control. 
UAVs operate in a continuous trajectory-based motion framework, where the desired trajectory is defined in the inertial frame. 
The UAV’s trajectory is parameterized using a waypoint trajectory generator (WTG), which provides a time-dependent reference position. The state of the UAV at each time step $t$ is given by:
$p_t = p_0 + \int_{0}^{t} v(\tau) \, d\tau$, 
where $p_t$ is the position at time $t$, $p_0$ is the initial position, and $v(t)$ is the velocity.

Physical attacks induce biases $b_t$ in sensor measurements, causing the control policy $\pi(\cdot)$ to generate unsafe control actions $a_t$ that may cause the UAV’s true state $o_t$ to deviate from the reference trajectory $g(t)$. 
This deviation is quantified as: $\Delta p_t = \| p_t - g(t) \| \gg \epsilon$, where $\epsilon$ defines a safety threshold, typically modeled as a circular region of radius $\epsilon$ centered around the target state $g(t)$. 
A trajectory is considered \emph{safe} if the UAV remains within this bound for all times, i.e., $\Delta p_t \leq \epsilon \; \forall t$; otherwise, it is considered \emph{unsafe}. 


%% file: sections/4-design.tex
The objective of \armor is to control UAVs in both adversarial and non-adversarial scenarios. 
An overview of our two-staged training approach is shown in Figure~\ref{fig:design}.

First, we train a teacher encoder that has access to privileged information ($X_t$), such as target sensor, manipulated states, attack-induced offset in physical states, and the duration of the attack.
The teacher encoder is based on variational autoencoder (VAE)~\cite{vae}, which receives both the UAV's states $O_t$ and $X_t$, and computes a latent embedding $\Bar{l_t}$ that represents the UAV's current state.
Next, we train a control policy using reinforcement learning, with the teacher encoder's latent embedding ($\bar{l_t}$) as input,
enabling the control policy to quickly learn and adapt to attack-induced state manipulations, and output resilient actions. 

Second, to enable real-world deployment where privileged information is unavailable, we train a \emph{student encoder} that relies solely on the onboard sensors. 
The student encoder is implemented as a temporal variational autoencoder (TVAE), that receives a sequence of historic physical state information ($H$) derived from onboard sensors. 
It computes a latent embedding $l_t$ in a supervised manner as shown in Figure~\ref{fig:design}, that approximates the teacher encoder's latent representation $\bar{l}_t$, enabling the same RL policy to operate reliably in the absence of privileged information. 

Our approach adopts a privileged learning strategy inspired by Lee \etal~\cite{priviledged-info}, but introduces two key innovations that improve both adversarial robustness and deployment efficiency.  
(1) We use the teacher encoder to generate a robust latent representation that is resilient to attack-induced perturbations, and the latent representation is the input to the control policy. This encourages the policy to rely entirely on a resilient representation, enhancing robustness to sensor manipulation. 
In contrast, Lee \etal combine raw observations with latent features, which can dilute the benefits of the robust representation.
(2) Rather than training separate teacher and student policies~\cite{priviledged-info}, we reuse a single control policy across both training and deployment. This eliminates the need to learn a second policy from scratch, simplifying the training pipeline and improving sample efficiency. 

\subsection{Stage I – Train with Privileged Information}
\label{sec:stage1}

We formulate the control problem as a Markov Decision Process (MDP) \cite{sutton2018ReinforcementLearning}. An MDP is defined by the tuple $(\mathcal{S}, \mathcal{A}, \mathcal{T}, \mathcal{R})$, where $\mathcal{S}$ is the state space, $\mathcal{A}$ is the action space, $\mathcal{T}$ is the transition probability $P(s_{t+1}|s_t, a_t)$, and $\mathcal{R}$ is a scalar reward function. 
The objective of the training framework is to learn a control policy $\pi(a_t | s_t)$ that maximizes the expected discounted sum of rewards over time.

In the teacher training stage of \armor, we assume a \emph{fully observable simulation environment}. The teacher encoder has access to both the UAV’s onboard sensor readings and privileged information that is not available during real-world deployment. 
The full state is defined as $s_t := \langle o_t, x_t \rangle$, where: $o_t$ includes the observable physical state of the UAV, such as position, angular orientation, heading direction, acceleration, linear and angular velocities. $x_t$ contains \emph{privileged information}, including the sensor under attack, the corresponding corrupted physical states, the magnitude of injected bias, and the duration of the attack. This information is extracted from the simulator and is used only during training.
The control action $a_t$ specifies low-level control targets for the UAV. 
Table~\ref{tab:input} summarizes the inputs and outputs of \armor. 

\begin{table}[!ht]
\footnotesize
\centering
\caption{\armor's inputs in both the training stages. $O_t$: physical states, $X_t$: Priviledged information, $S_t$: Inputs to teacher encoder, $H$: Inputs to student encoder, and $a_t$: action. }
\label{tab:input}
\begin{tabular}{c|l}
\hline
\textbf{Input Type} & \textbf{Description}                                                                                                                                                \\ \hline
$O_t$               & \begin{tabular}[c]{@{}l@{}}position, velocity, orientation, body angular rates.\\ $[x, y, z, \dot{x}, \dot{y}, \dot{z}, \phi, \theta, \psi, \dot{\phi}, \dot{\theta}, \dot{\psi}]$,\end{tabular} \\ \hline
$X_t$               & \begin{tabular}[c]{@{}l@{}}[target sensor, corrupted physical states, \\ bias intensity, duration]\\ example: [GPS, ($x, y, z$), (-5, 0, 0), 10]\end{tabular}       \\ \hline
$S_t$               & $S_t = \langle O_t, X_t \rangle$                                                                                                                 \\ \hline
$H$                 & $H = \{ o_{t-n}, .. , o_{t-3}, o_{t-2}, o_{t-1} \}$,                                                                                                                \\ \hline
$a_t$               & Position and attitude control commands - $x, y, z$ axes.                                                                                                            \\ \hline
\end{tabular}
\end{table}

The teacher encoder is implemented as a multi-head variational autoencoder (VAE) that maps the input $s_t = \langle o_t, x_t \rangle$ to three outputs:
$f_{\text{teacher}}(s_t) = \left( \bar{l}_t = (\mu_t, \sigma_t), \; \hat{y}_t, \; \hat{s}_t \right)$, 
where $\bar{l}_t$ is the latent representation with mean $\mu_t$ and variance $\sigma_t$, $\hat{y}_t$ is the predicted attack type, and $\hat{s}_t$ is the reconstruction of the input $s_t$. 
\armor leverages both the $\mu_t$ and $\sigma_t$ of the latent representation, allowing the policy to adjust control when uncertainty is increases ($\sigma_t\uparrow$), such as under stealthy attacks~\cite{stealthy-attacks}. 
The attack-type classifier encourages the latent space to capture attack-specific patterns, while the variance $\sigma_t$ provides an uncertainty estimate that allows the control policy $\pi(a_t|\bar{l}_t)$ to adjust its behavior under unseen attacks.

The teacher encoder is trained with an auxiliary decoder to evaluate the quality of the latent representations through reconstruction loss; however, the decoder is discarded after training, and only the encoder is used with the RL policy. 
The teacher encoder is trained by minimizing a combined loss that includes the reconstruction loss, the Kullback-Leibler (KL) divergence, and the attack classification loss:

\begin{equation}
\thinspace
    \mathcal{L}_{\text{teacher}} = \mathcal{L}_{\text{recon}} + \mathcal{L}_{\text{KL}} + \mathcal{L}_{\text{attack}}
\end{equation}

The RL policy $\pi(a_t | \bar{l}_t)$ is trained using Proximal Policy Optimization (PPO)~\cite{ppo}, with the latent representation $\bar{l}_t$ as input. This design encourages the policy to rely on a robust, attack-aware latent representation, improving its resilience.

This stage enables the policy to exploit privileged information during training, allowing it to generate \emph{resilient behaviors} under adversarial conditions. The resulting latent space is a robust representation for policy learning and student encoder supervision in Stage II.

The \emph{reward function} is designed to promote task completion while ensuring safety and stability. For instance, the RL agent receives positive reward for minimizing distance to the target waypoint and penalties for unsafe behaviors such as abrupt motion, excessive tilt, or deviation from trajectory. 
The reward function is defined as:

\begin{equation}
\label{eqn:reward-func}
\begin{split}
r_t = \; & R_{\text{goal}} \cdot \exp\left(-\lambda \cdot \|p_t - g_t\|\right) - \alpha \cdot \|p_t - p_{t-1}\| - \beta \cdot \theta_t \\
& - \gamma \cdot \|a_t - a_{t-1}\|^2
\end{split}
\end{equation}

where $p_t$ is the UAV's position, $g_t$ is the goal waypoint, and $\|p_t - g_t\|$ is the Euclidean distance to the target. 
The exponential term provides a smooth approximation of the goal reward, sharply increasing as the UAV nears the goal. 
The remaining terms penalize trajectory deviation, tilt ($\theta_t$), and abrupt control actions ($a_t$), with corresponding weights $\alpha$, $\beta$, and $\gamma$.
The term $\theta_t$ represents the total tilt of the UAV (e.g., combined roll and pitch deviation), and $a_t$ represents the control command at time $t$ (e.g., change of position). 
The coefficients $\alpha$, $\beta$, and $\gamma$ weight the penalties for deviation from the goal, instability, and abrupt motion, respectively.


\subsection{Stage II – Transfer Learning Adaptation}
\label{sec:stage2}

In this stage, we introduce a \emph{student encoder} that operates solely on data derived from onboard sensors. The core idea is to approximate the teacher's latent representation $\bar{l}_t$ using only historical UAV physical state. 
This enables a transfer learning setup in which the RL policy, originally trained with privileged information, can now operate with latent representations generated by the student encoder.

We implement the student encoder as a temporal variational autoencoder (TVAE) built using a Long Short-Term Memory (LSTM) network, which effectively models sequential dependencies in time-series sensor data. 
The encoder takes as input a sequence of historical UAV physical states in a sliding window $H := \{o_{t-N}, \dots, o_{t-1}\}$, where each $o_t$ denotes the UAV’s physical state at time $t$ (e.g., position, velocity, angular rate, and orientation). 
This sequence $H$ provides temporal context that implicitly captures the impact of adversarial disturbances over time. 
The student encoder maps $H$ to a latent representation $l_t$ that approximates the teacher encoder’s latent representation $\bar{l}_t$. 
$l_t$ is then passed to the trained control policy to derive the control actions. 

The student encoder outputs: (i) a latent representation $l_t$ (mean and variance), (ii) an attack-type prediction, and (iii) a reconstruction of the input (during training). 
This multi-head design encourages the latent space to separate task-relevant features from attack-induced perturbations, improving generalization across different attack types. 

The student encoder is trained via supervised learning, using the teacher encoder's outputs as targets. For each input history $H_t$, the student aims to approximate the teacher’s latent representation $\bar{l}_t$ and ensure that the control policy produces consistent actions from both representations. Specifically, we minimize the combined loss:

\begin{equation}
\thinspace
    \mathcal{L}_{\text{student}} := \| l_t(H_t) - \bar{l}_t(o_t, x_t) \|^2 + \| a_t(l_t) - a_t(\bar{l}_t) \|^2 + \mathcal{L}_{\text{attack}}
\end{equation}
   
where the first term encourages the student encoder to match the teacher’s latent representation, the second term aligns policy outputs, and $\mathcal{L}_{\text{attack}}$ penalizes errors in attack-type classification.
Once training is complete, the RL control policy originally trained with $\bar{l}_t$ is reused and unchanged. At deployment, the RL control policy takes $l_t$ as input, enabling attack-resilient control using only UAV onboard sensor data.

%% file: sections/5-results.tex
In this section, we first outline the experimental setup, the simulation environment, and the metrics used for evaluation. 
Then, we present results evaluating the effectiveness of \armor across three key aspects: 
(1) The performance of the student encoder in approximating the teacher’s latent state representations without access to privileged information. 
(2) The ability to maintain safe and stable flight under physical attacks.
(3) Generalization to unseen attack scenarios. 

{\em \textbf{Physical Attacks.}}
We evaluate \armor in the presence of {\em five different types of physical attacks~\cite{physical-attacks}} targeting the GPS, gyroscope, accelerometer, magnetometer, and optical flow sensors of the UAV. 
We simulate realistic physical attacks using RAVAGE~\cite{ravage}, a software tool that supports launching realistic physical attack signals (attack bias, attack duration, bias pattern). 
Table~\ref{tab:attack-params} outlines the attack parameters. 

As summarized in the table, the attacks differ in nature, ranging from drift to oscillatory and random bias patterns, with varying intensity and attack duration. 
This represents different manipulation strategies across sensor modalities, ensuring that evaluation covers a broad spectrum of adversarial conditions rather than variations of a single attack type.
In addition, we evaluate under stealthy attacks, where small biases accumulate gradually over time to destabilize the UAV.

\begin{table}[!ht]
\centering
\footnotesize
\caption{Attack types for evaluation. Attack parameters - intensities, bias patterns, and duration.}
\label{tab:attack-params}
\begin{tabular}{l|l|c|c}
\hline
\multirow{2}{*}{\textbf{Sensors}} & \multirow{2}{*}{\textbf{Bias Type}} & \multirow{2}{*}{\textbf{Bias Range}} & \multirow{2}{*}{\textbf{Attack Duration}} \\
                                        &                                       &                                      &                                           \\ \hline
GPS                                     & drift                                 & 1-20 $m$                               & up to 60s                                 \\ 
Gyroscope                               & oscillatory                           & 1-90 $deg$                             & up to 60s                                 \\ 
Accelerometer                           & oscillatory                           & 0.5-1 $m/s^2$                          & up to 30s                                 \\ 
Magnetometer                            & random                                & 10-90 $deg$                            & up to 60s                                 \\ 
Optical Flow                            & random                                & 0.1-0.5 $m/s$                          & up to 30s                                 \\ \hline
\end{tabular}
\end{table}

{\em \textbf{Simulation Environment.}}
We consider a quadcopter operating in 3D space. 
The UAV dynamics are simulated using \texttt{gym-pybullet}~\cite{gympybullet}, an OpenAI Gym-compatible~\cite{openai-gym} environment built on the PyBullet physics engine, which provides a realistic simulation of rigid-body dynamics. 
As in prior work~\cite{safe-control-gym}, we use standard quadcopter equations of motion with translational and rotational dynamics.
We collected privileged information for training using the RAVAGE~\cite{ravage} tool for injecting attacks into UAVs. 
Note that, due to space constraints, we focus on a single vehicle model and present comprehensive results, rather than exploring multiple testbeds and sim-to-real deployment. 

{\em \textbf{Control Objective}}. The control objective is to reach a randomly sampled goal position $g$ in 3D space, represented as a sphere with a radius of $0.1\,\text{m}$. 
To guide the UAV towards the goal while ensuring stable and safe flight, we design a shaped reward function that provides incentives for reaching the goal and penalties for unsafe behaviors (Equation~\ref{eqn:reward-func}). 

{\em \textbf{Implementation.}}
Both the teacher and student encoders use a latent vector size of 20. 
The student encoder takes a historic sequence of 200 timesteps as input. 
The RL control policy is trained using PPO~\cite{ppo}  (Section~\ref{sec:stage1}). 

{\em \textbf{Comparison.}} 
We compare the effectiveness of \armor with two state-of-the-art approaches: (1) Robust Adversarial Reinforcement Learning (RARL)~\cite{rarl}, which formulates adversarial training as a minimax game between a protagonist and an adversary, aiming to learn a policy that is robust to sensor perturbations. 
(2) Hybrid Recovery Policy (HRP)~\cite{recovery-rl}, which uses an RL policy for control in unsafe zones and a stabilizing PID controller for safety outside unsafe zones. 



{\em \textbf{Ablation Study.}}
We implement a baseline-RL policy with the same architecture as \armor, except that it does not incorporate any encoder. 
This baseline directly processes high-dimensional physical state information as input to the PPO policy, without mapping it into a latent representation. 
The baseline serves as an ablation to evaluate the effectiveness of encoders in improving resilience against physical attacks.

{\em \textbf{Evaluation Metrics.}}
We use the following three metrics: 

\begin{enumerate}
    \item \textbf{Mission Success Rate} measures the proportion of episodes in which the UAV successfully reaches the goal position $g$ within an error margin $\epsilon$. 
    A mission is considered successful if the final UAV position $p_T$ satisfies $\|p_T - g\| \leq \epsilon$, where $\epsilon=5m$~\cite{specguard, pid-piper}.
    \item \textbf{Crash Rate} measures the proportion of episodes in which the mission failed due to a crash. A crash is defined as the UAV's state exceeding predefined safety bounds, resulting in termination of the episode.
    \item \textbf{State Drift} measures the mean absolute deviation in the physical states from the ideal physical states during attacks. For example, in the case of GPS attacks, state drift is quantified as the Euclidean distance between the UAV's current position $p_t$ and the ideal position $\hat{p}_t$ at each time $t$ during the attack duration $T$.  
\end{enumerate}

\subsection{\armor Training}
\label{sec:res-training}

\begin{figure}[!ht]
  \centering
  \begin{subfigure}{0.23\textwidth}
    \includegraphics[width=\textwidth]{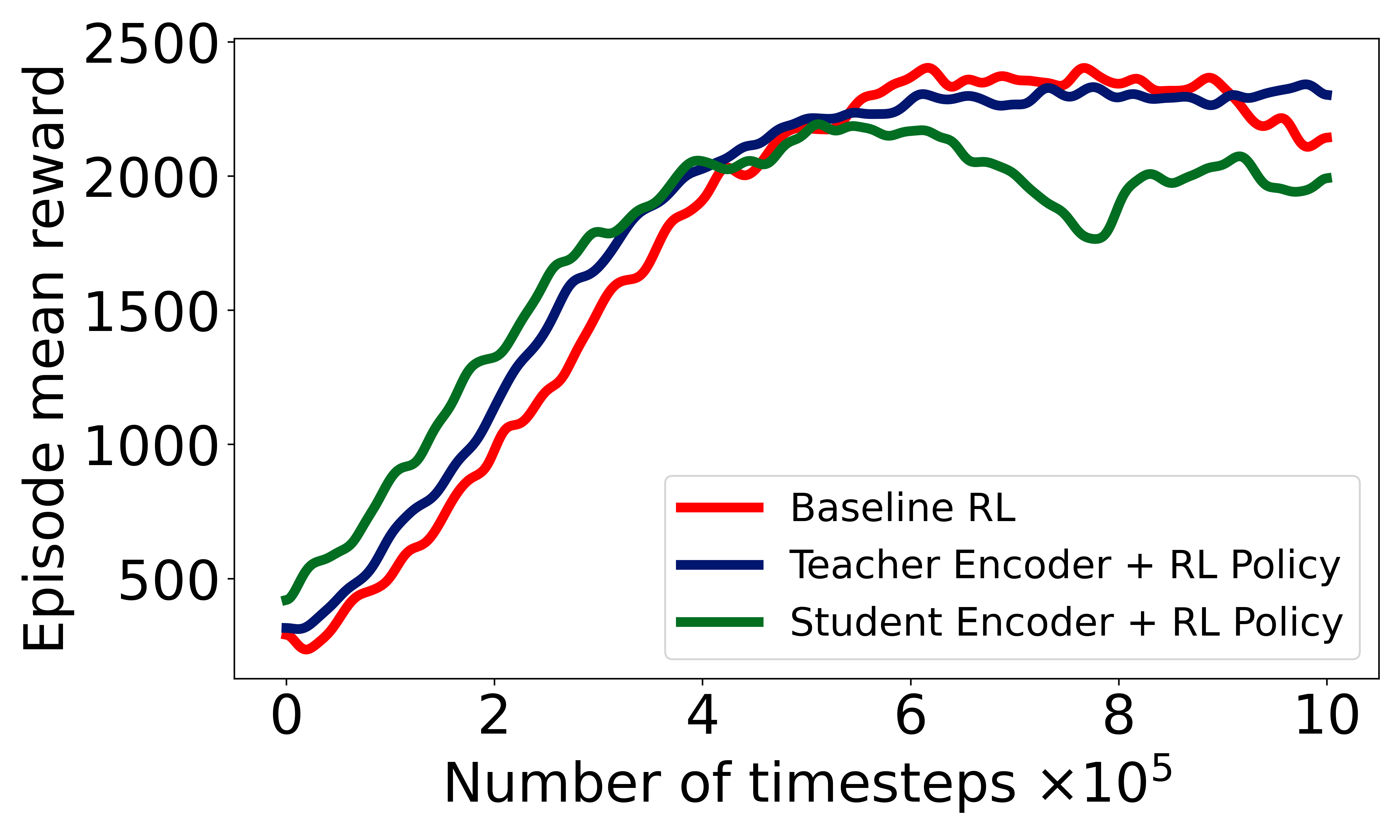}
  \end{subfigure}
  \begin{subfigure}{0.23\textwidth}
    \includegraphics[width=\textwidth]{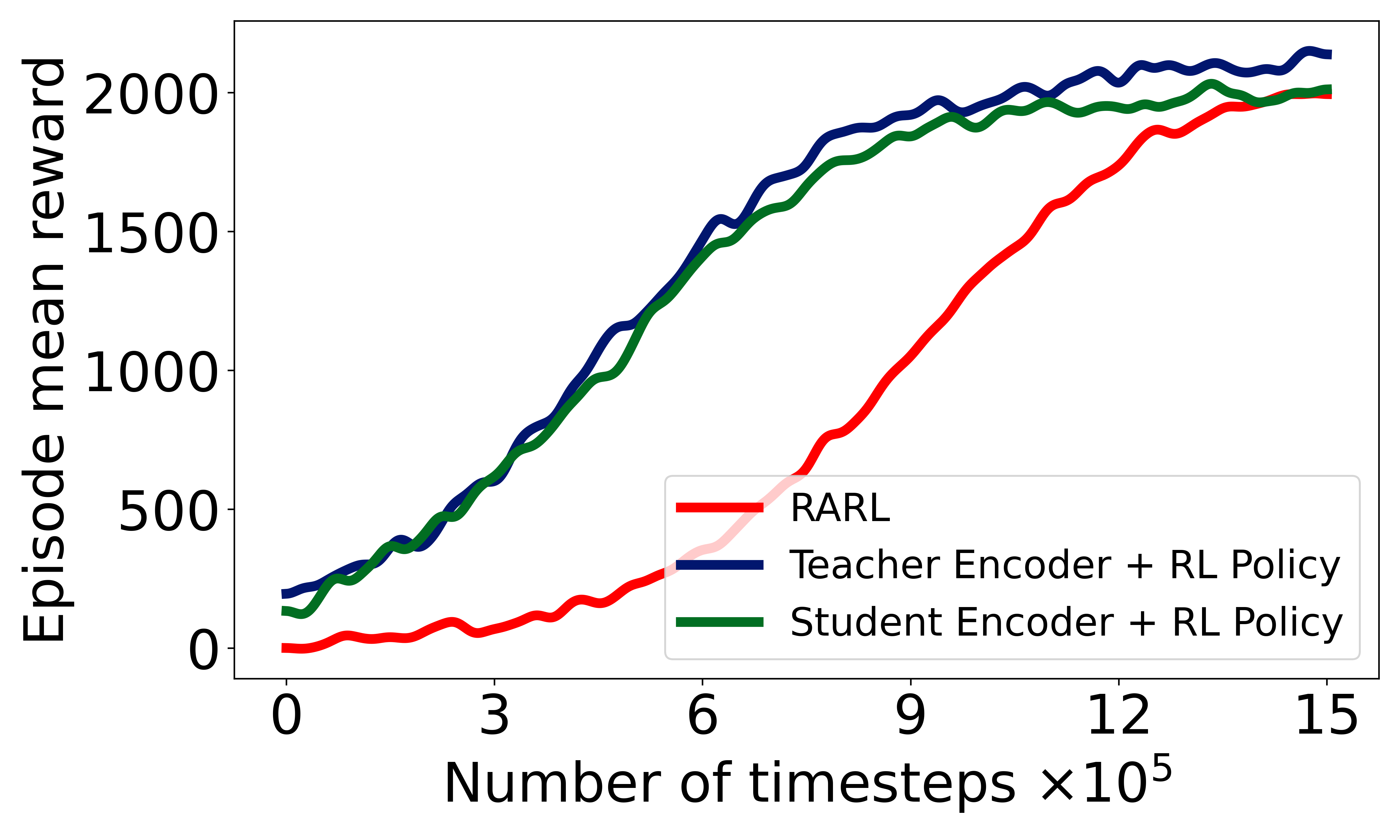}
  \end{subfigure}
    \caption{
    Training performance comparison. 
    \textbf{Left:} Nominal conditions, all methods achieve similar final performance. 
    \textbf{Right:} Adversarial conditions, both the Teacher and Student encoder policies significantly accelerate learning compared to RARL. The Student Encoder policy enables faster convergence even without access to privileged information. 
    }
  \label{fig:training-comparison}
\end{figure}

Figure~\ref{fig:training-comparison} compares the training performance of \armor in two scenarios: (a) no-attack conditions (nominal conditions), where we compare \armor with baseline-RL, and (b) adversarial conditions (physical attacks), where we compare \armor with an adversarially trained control policy.
We refer to the two variants of \armor - the Teacher Encoder policy and the Student Encoder policy as the RL control policies that use latent representations from the Teacher and Student Encoders, respectively.
The figures represent the mean performance averaged over 5 random seeds. 

As shown in Figure~\ref{fig:training-comparison}(a), in the absence of attacks, all the methods: baseline-RL, the Teacher Encoder policy, and the Student Encoder policy, achieve comparable final episodic rewards, converging to approximately 2200 after 500k timesteps. 
The similarity in final performance indicates that the use of encoder-based latent representations in the Teacher and Student encoders does not hinder the ability of the policy to learn optimal control under nominal conditions. 

Figure~\ref{fig:training-comparison}(b) shows the results in adversarial conditions. 
We train a control policy using the RARL~\cite{rarl} approach for adversarial robustness. This policy converges slowly, requiring nearly 1200k timesteps, twice as long, to reach a reward of 2100.
In contrast, both the Teacher Encoder policy and the Student Encoder policy demonstrate significantly faster convergence speed (2$\times$ faster). 
The Teacher Encoder policy, trained with privileged information, reaches a reward of around 2100 in under 600k timesteps. 
Notably, the Student Encoder policy, despite lacking access to privileged information, also achieves similar convergence. 
{\em These results demonstrate that the \armor's two-stage training is effective in transferring robustness from the attack-aware privileged learning phase to the online deployment phase.} 

Henceforth, we refer to the Student Encoder policy used for online inference as \armor. 
Under attack-free conditions, \armor matches the control performance of a baseline PID controller reported in prior work using the \texttt{gym-pybullet} environment~\cite{gympybullet}. This shows that \armor's robustness to physical attacks does not compromise nominal performance.

\subsection{Effectiveness of \armor under Physical Attacks}
\label{sec:res-armor-effectiveness}

\begin{figure}[htbp]
    \centering
    \begin{subfigure}[b]{0.22\linewidth}
        \includegraphics[width=\linewidth]{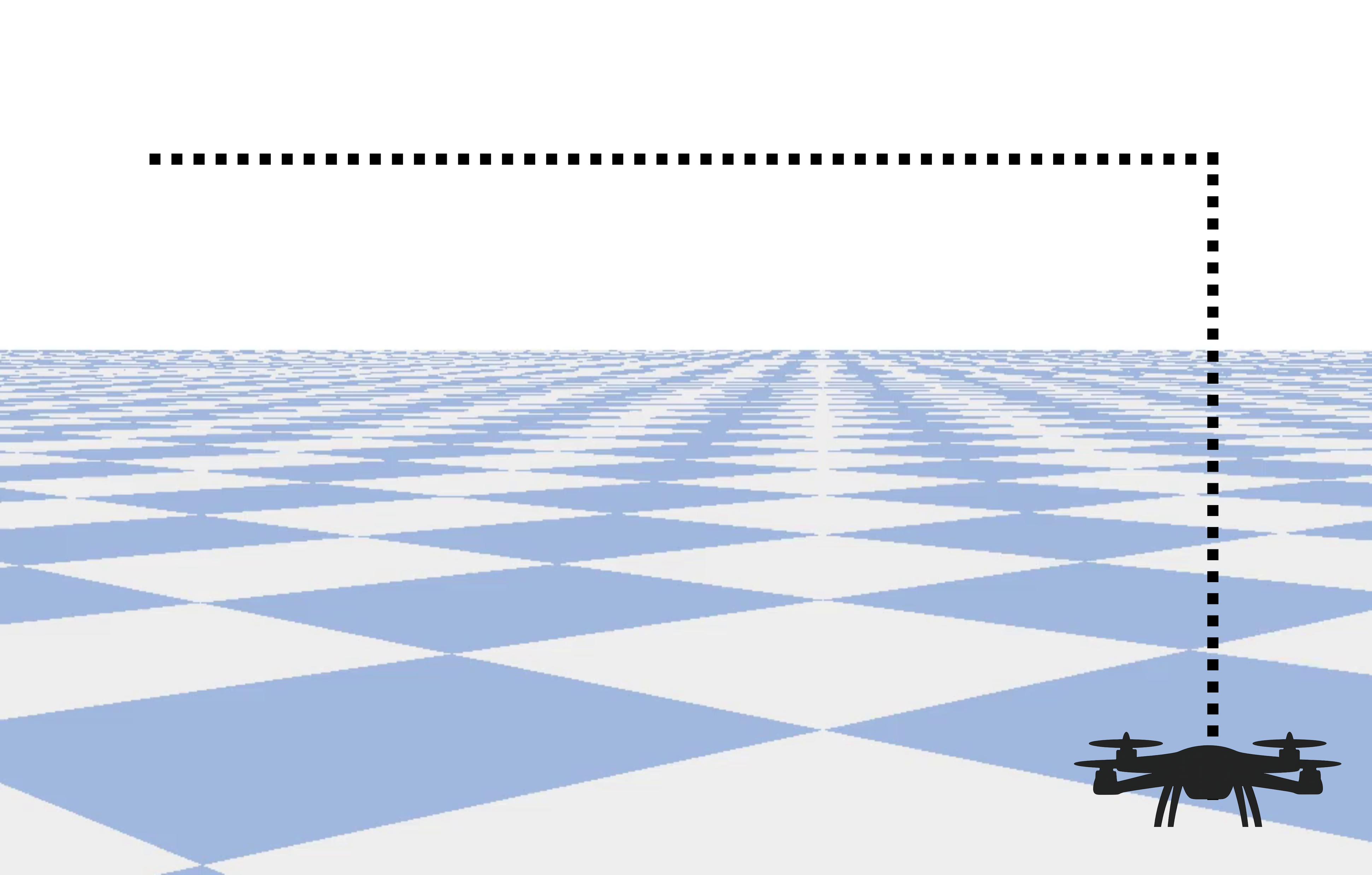}
    \end{subfigure}
    \begin{subfigure}[b]{0.22\linewidth}
        \includegraphics[width=\linewidth]{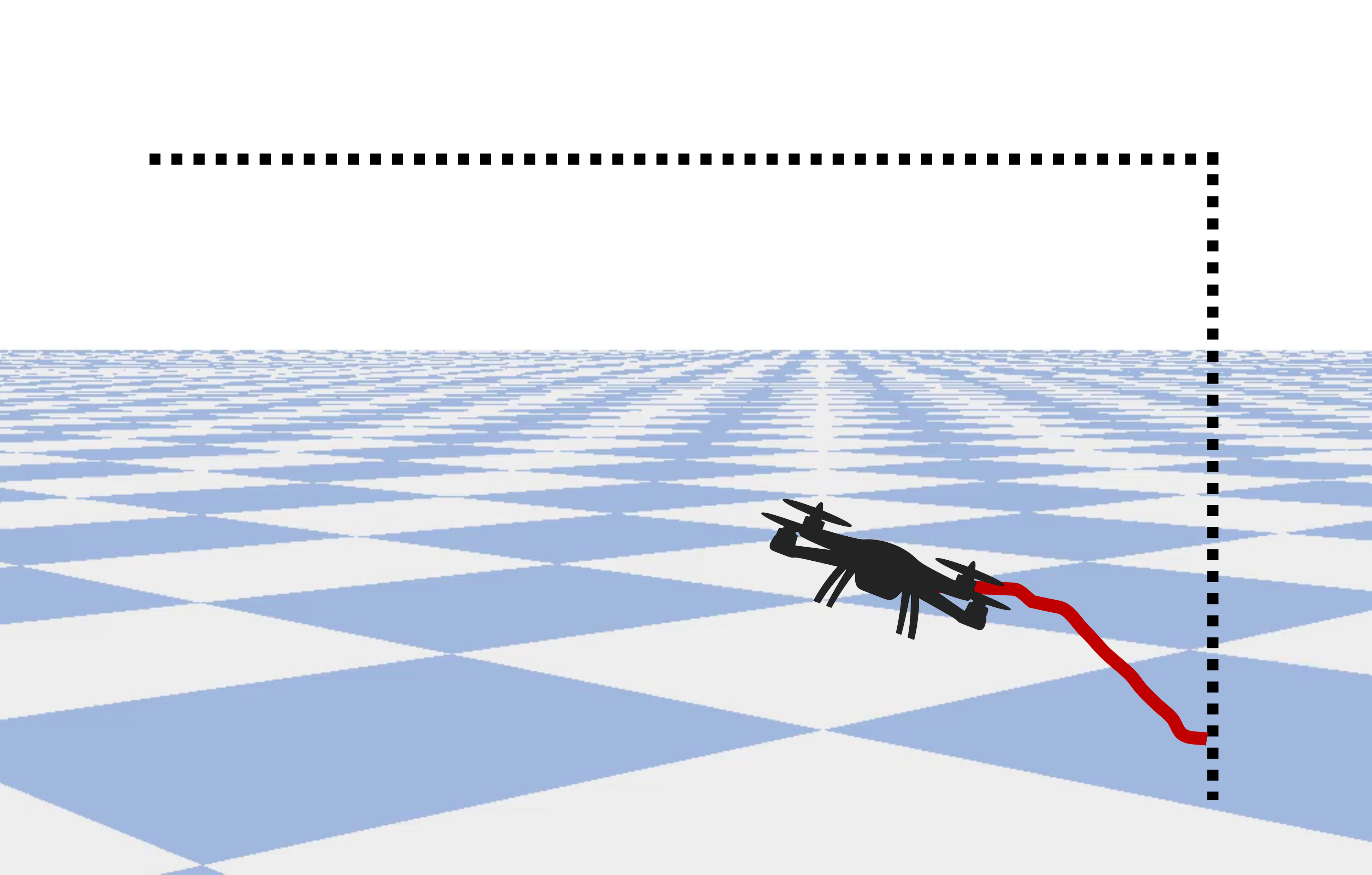}
    \end{subfigure}
    \begin{subfigure}[b]{0.22\linewidth}
        \includegraphics[width=\linewidth]{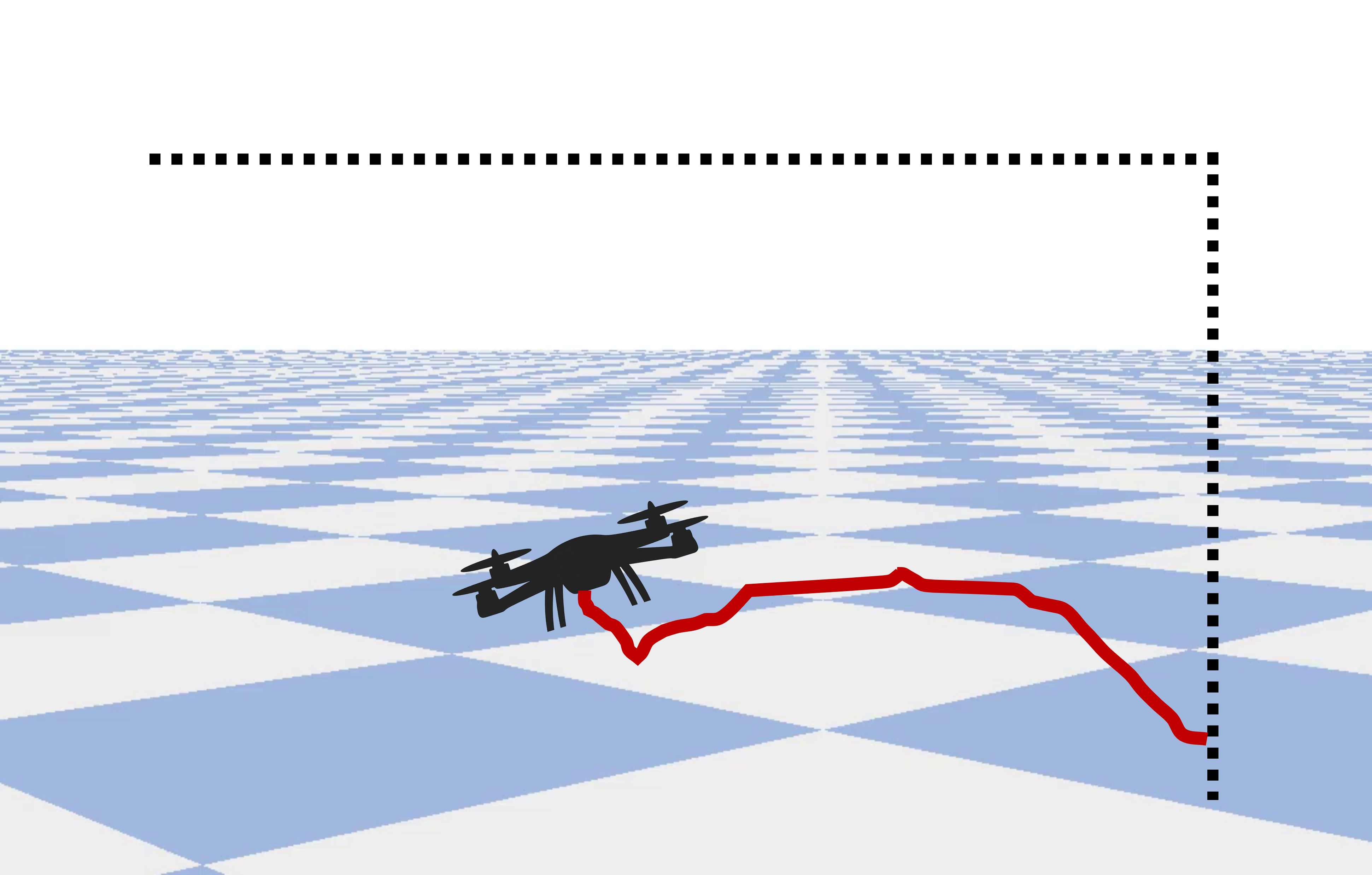}
    \end{subfigure}
    \begin{subfigure}[b]{0.22\linewidth}
        \includegraphics[width=\linewidth]{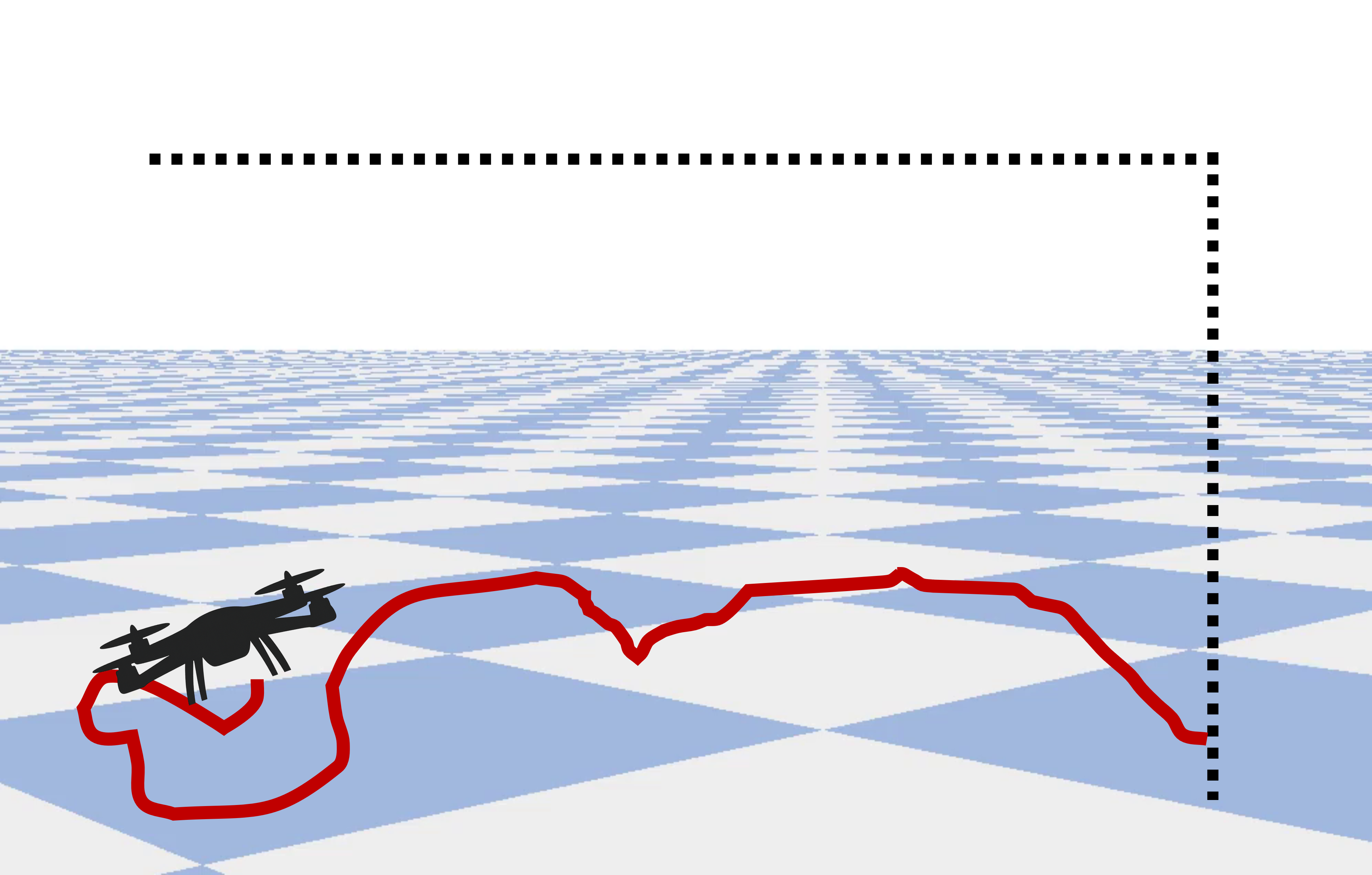}
    \end{subfigure}
    \begin{subfigure}[b]{0.22\linewidth}
        \includegraphics[width=\linewidth]{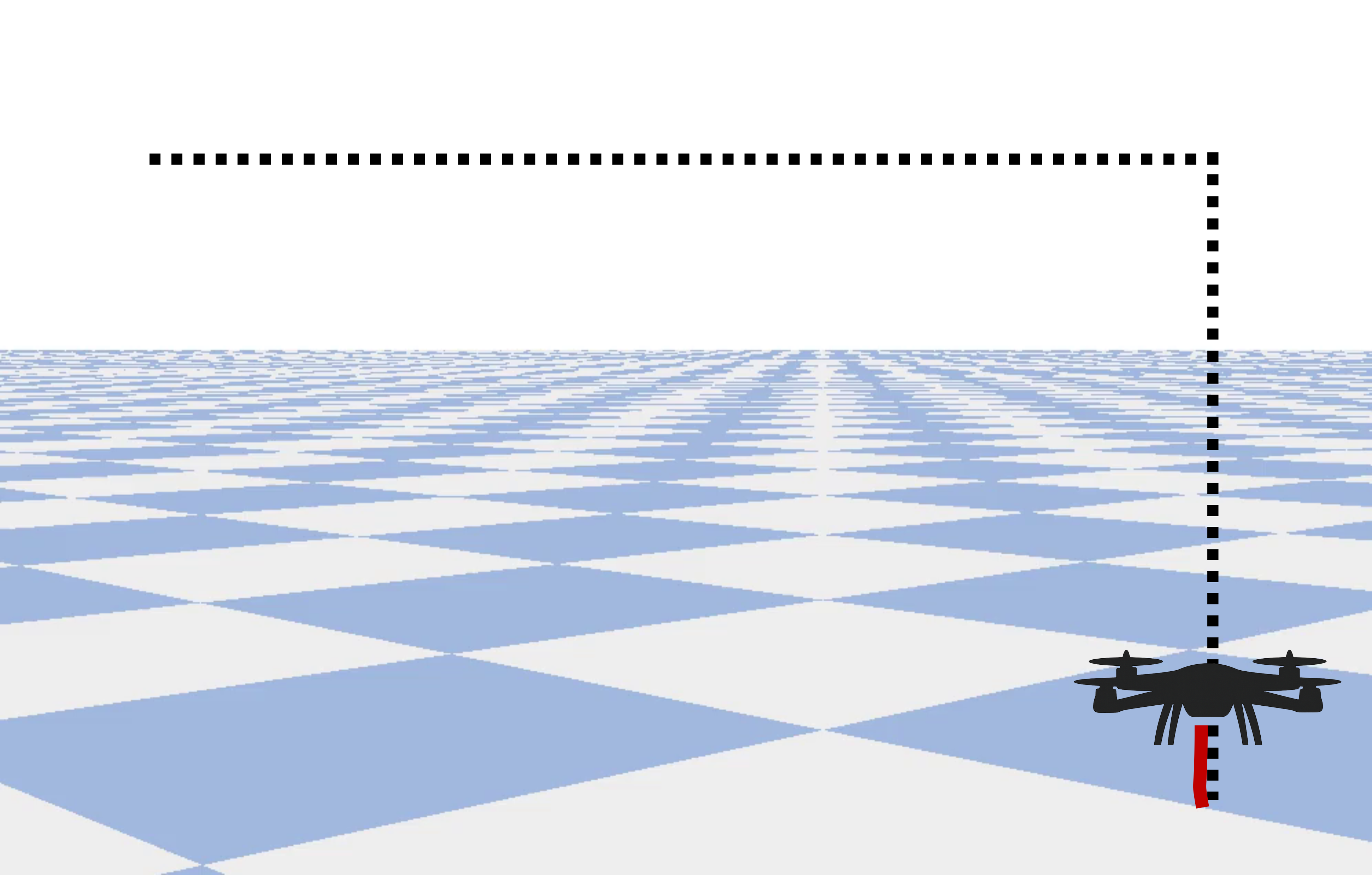}
    \end{subfigure}
    \begin{subfigure}[b]{0.22\linewidth}
        \includegraphics[width=\linewidth]{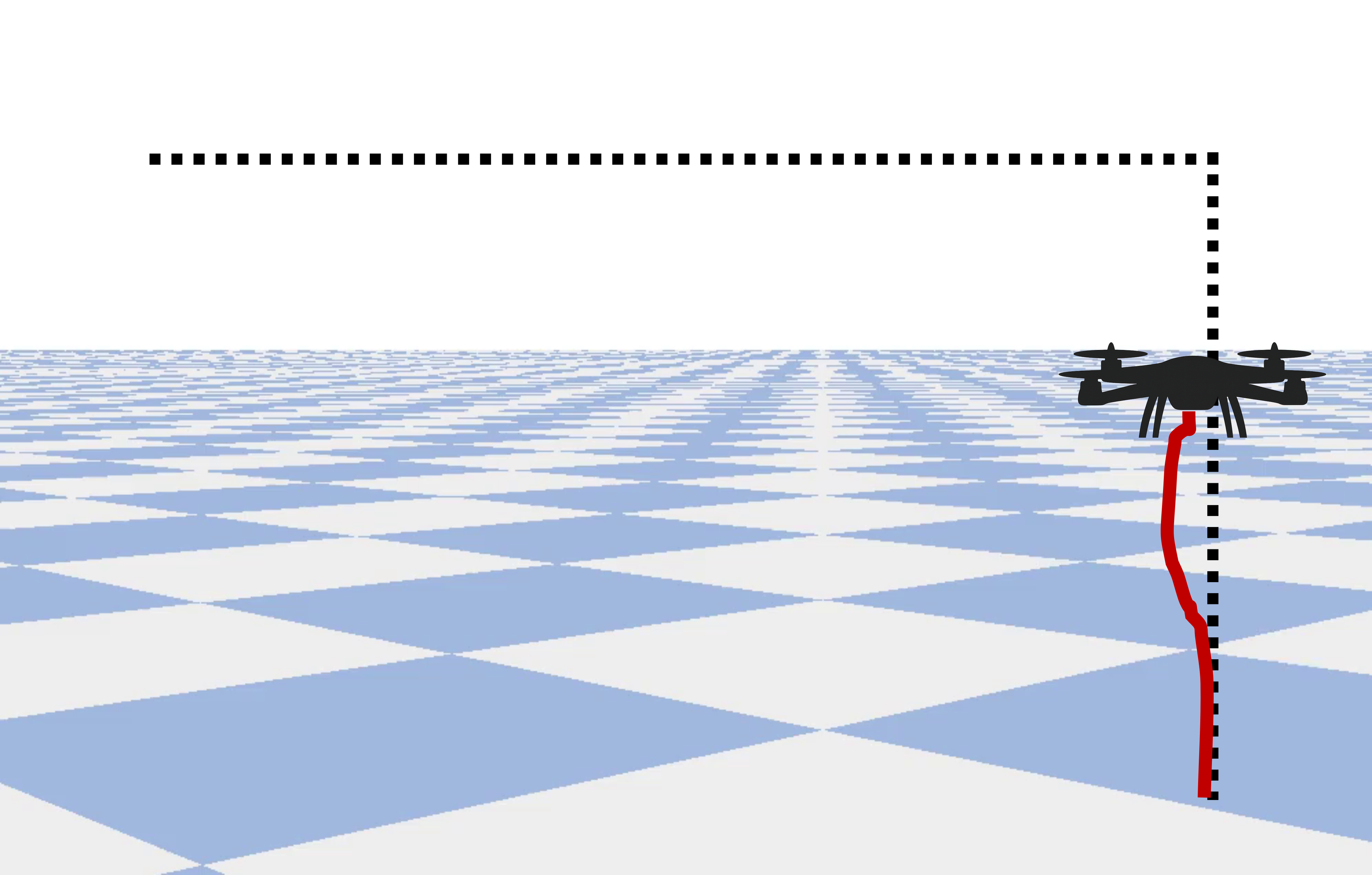}
    \end{subfigure}
    \begin{subfigure}[b]{0.22\linewidth}
        \includegraphics[width=\linewidth]{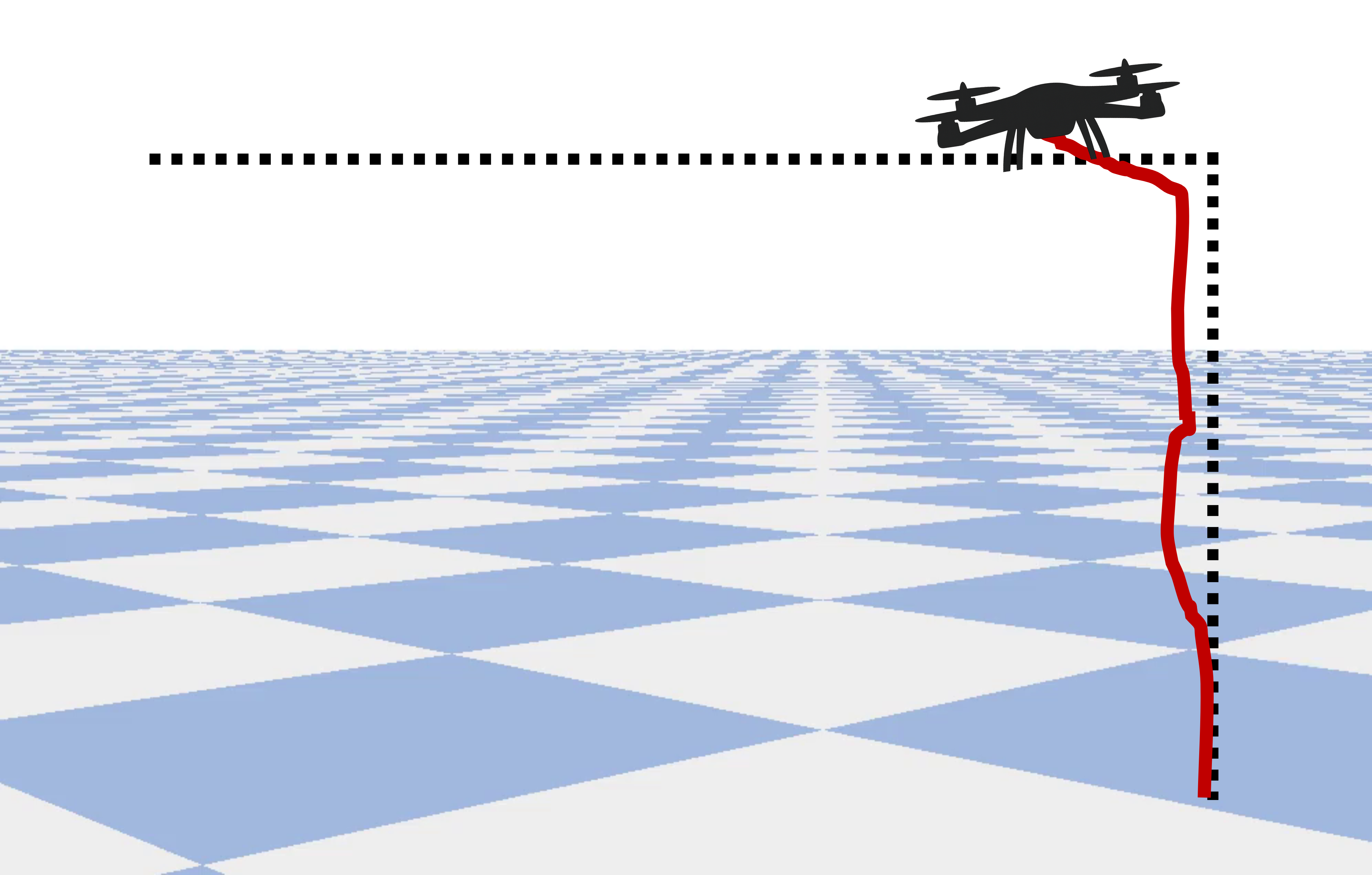}
    \end{subfigure}
    \begin{subfigure}[b]{0.22\linewidth}
        \includegraphics[width=\linewidth]{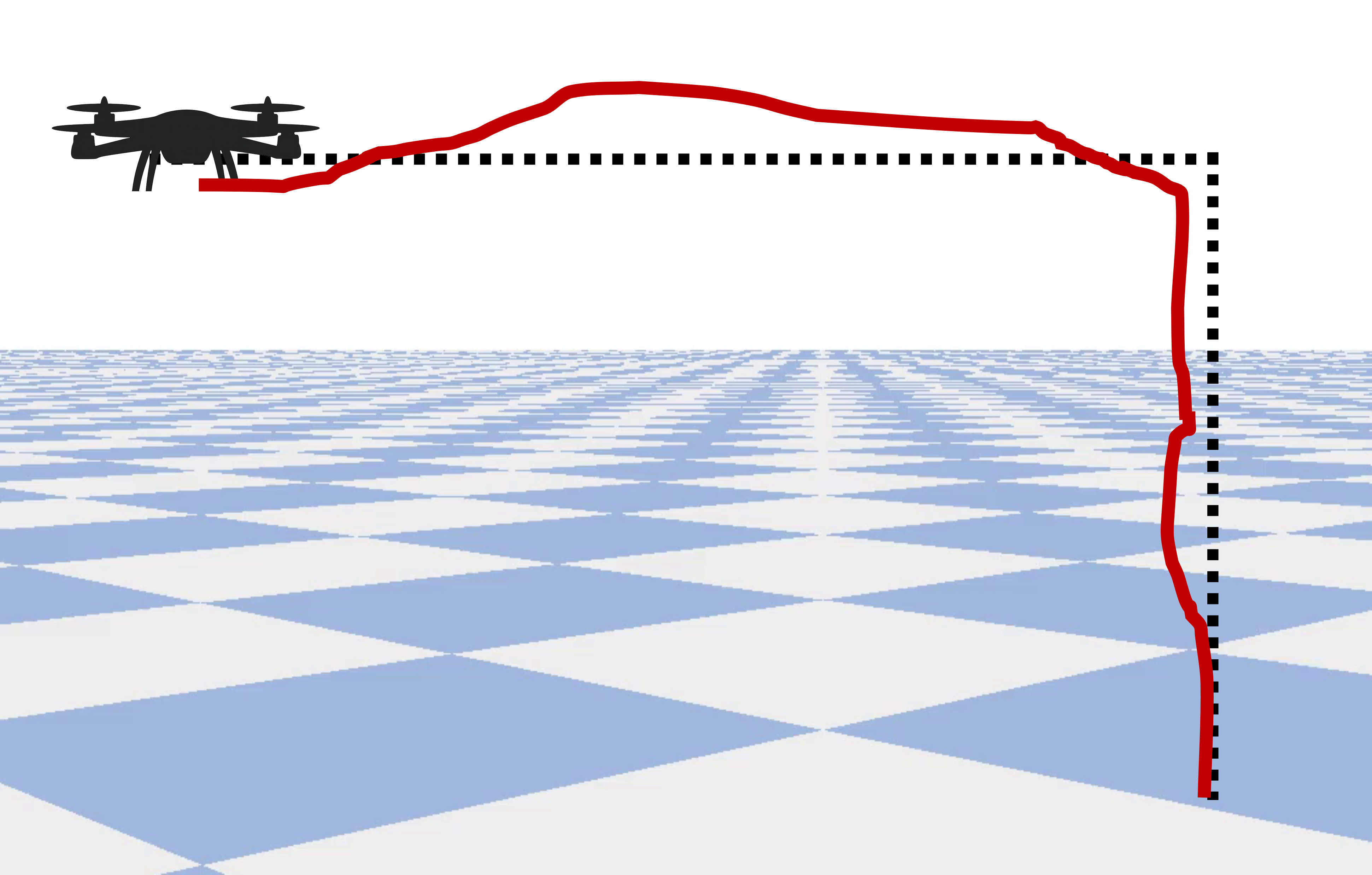}
    \end{subfigure}
    \caption{Control performance under GPS spoofing attack. The \textbf{Top row} shows the trajectory deviations with baseline-RL. The \textbf{Bottom row} shows the trajectory with \armor, demonstrating resilient control despite the attack.}
    \label{fig:armor-gps}
\end{figure}

Figure~\ref{fig:armor-gps} shows 
the effectiveness of \armor under GPS spoofing attacks. 
The red lines represent the UAV's actual trajectory. 
With baseline-RL (top row), the UAV deviates significantly from the intended path (in dotted line) due to incorrect position estimates, resulting in a crash. 
In contrast, with \armor (bottom row), the UAV maintains stable flight with minimal deviation from the intended path, successfully completing the mission despite the attack.

We evaluate \armor under five different types of physical attacks shown in Table~\ref{tab:rarl_armor}. 
\armor demonstrates strong resilience against all attack types, maintaining safe and stable flight. 
\armor achieves an average success rate of 88\%, incurring 0 crashes. 
Even when the missions failed, \armor prevented crashes and maintained minimal state drift. 

\subsection{Comparison with Baseline-RL, HRP and RARL}
\label{sec:res-armor-comparison}

First, we discuss two cases in detail comparing \armor with baseline-RL and HRP under two different attack types with different bias patterns: (1) GPS spoofing, which introduces drift biases in position estimates, and (2) gyroscope attack, which induces oscillatory biases in attitude estimates. We then present a more comprehensive comparison. 

\begin{figure}[!ht]
  \centering
  \begin{subfigure}{0.23\textwidth}
    \includegraphics[width=\textwidth]{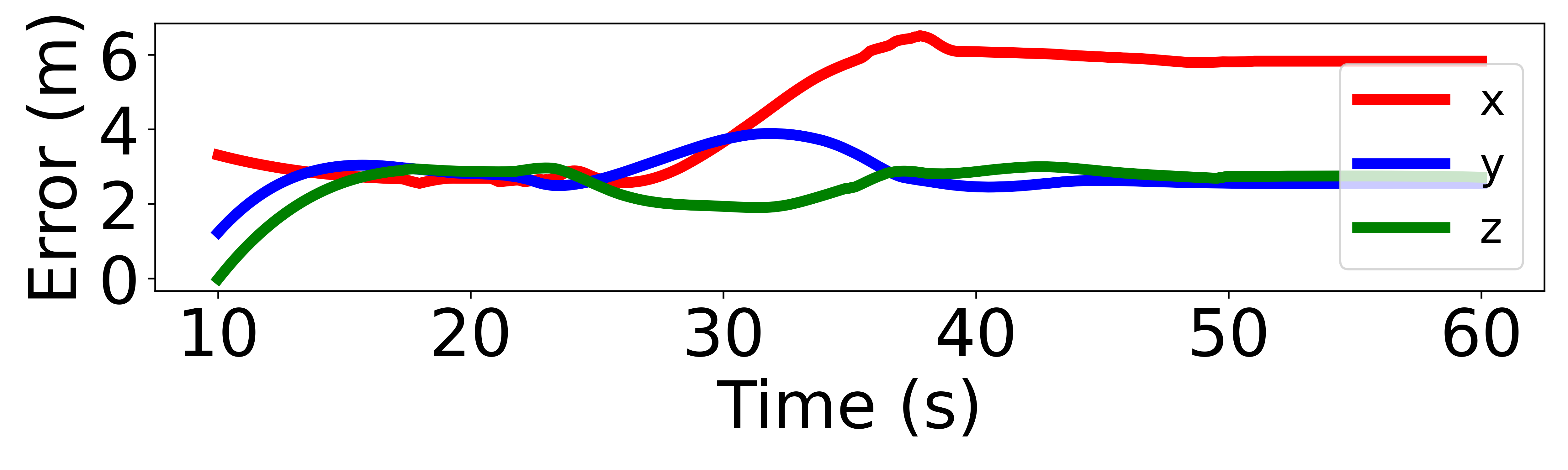}
  \end{subfigure}
  \begin{subfigure}{0.23\textwidth}
    \includegraphics[width=\textwidth]{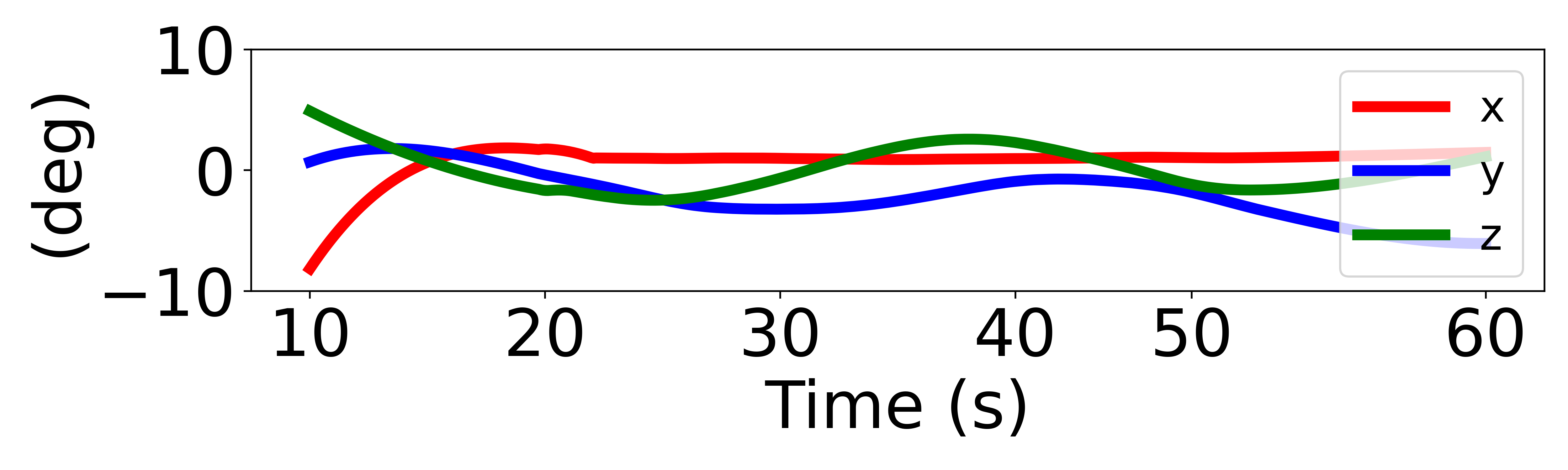}
  \end{subfigure}
  \begin{subfigure}{0.04\textwidth}
    \captionsetup{labelformat=empty}
  \end{subfigure}

  \begin{subfigure}{0.23\textwidth}
    \includegraphics[width=\textwidth]{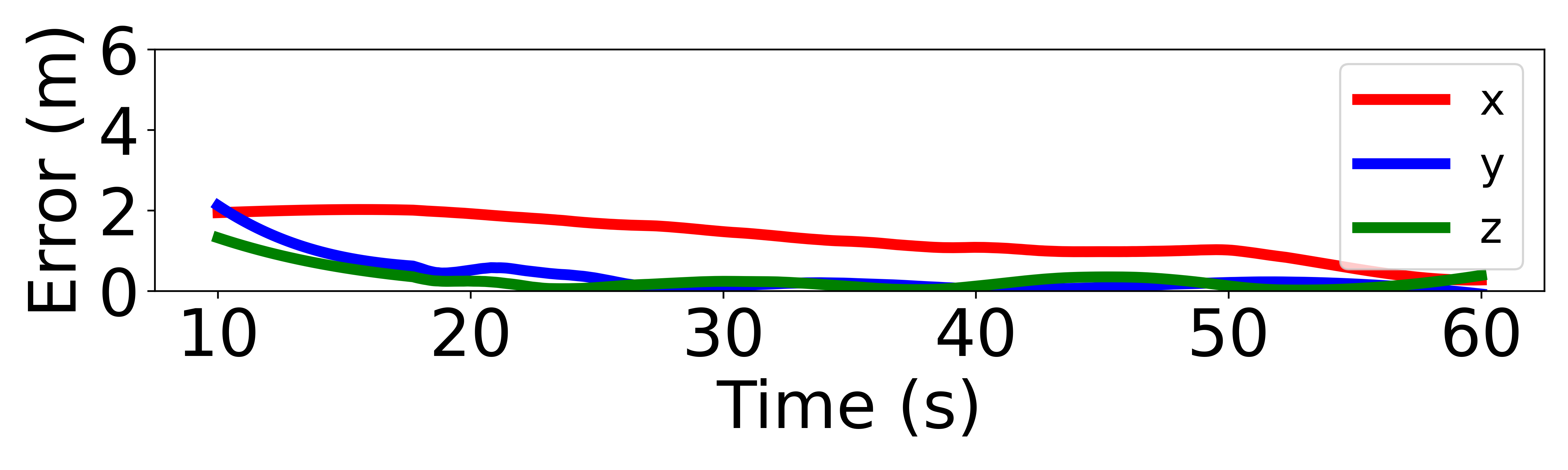}
  \end{subfigure}
  \begin{subfigure}{0.23\textwidth}
    \includegraphics[width=\textwidth]{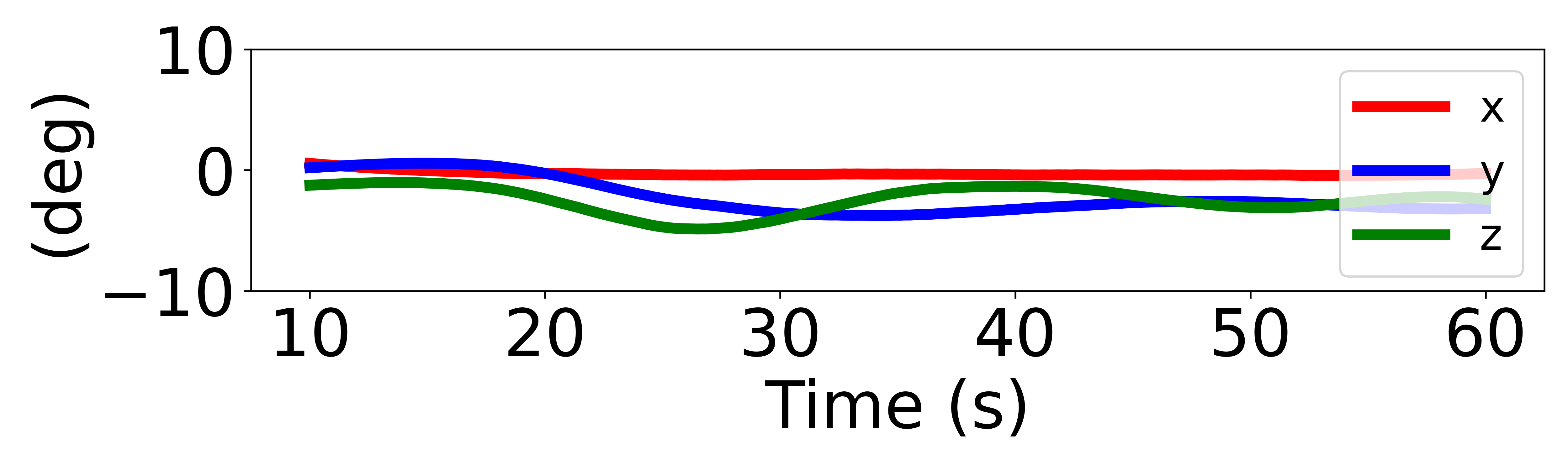}
  \end{subfigure}
  \begin{subfigure}{0.04\textwidth}
    \captionsetup{labelformat=empty}
  \end{subfigure}

  \begin{subfigure}{0.23\textwidth}
    \includegraphics[width=\textwidth]{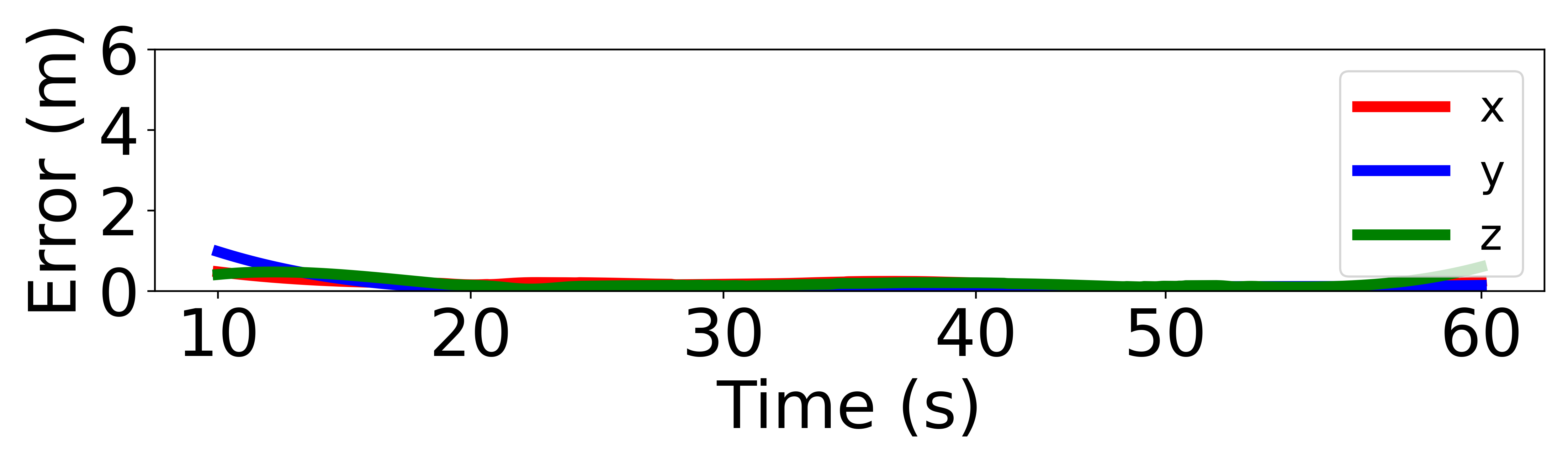}
  \end{subfigure}
  \begin{subfigure}{0.23\textwidth}
    \includegraphics[width=\textwidth]{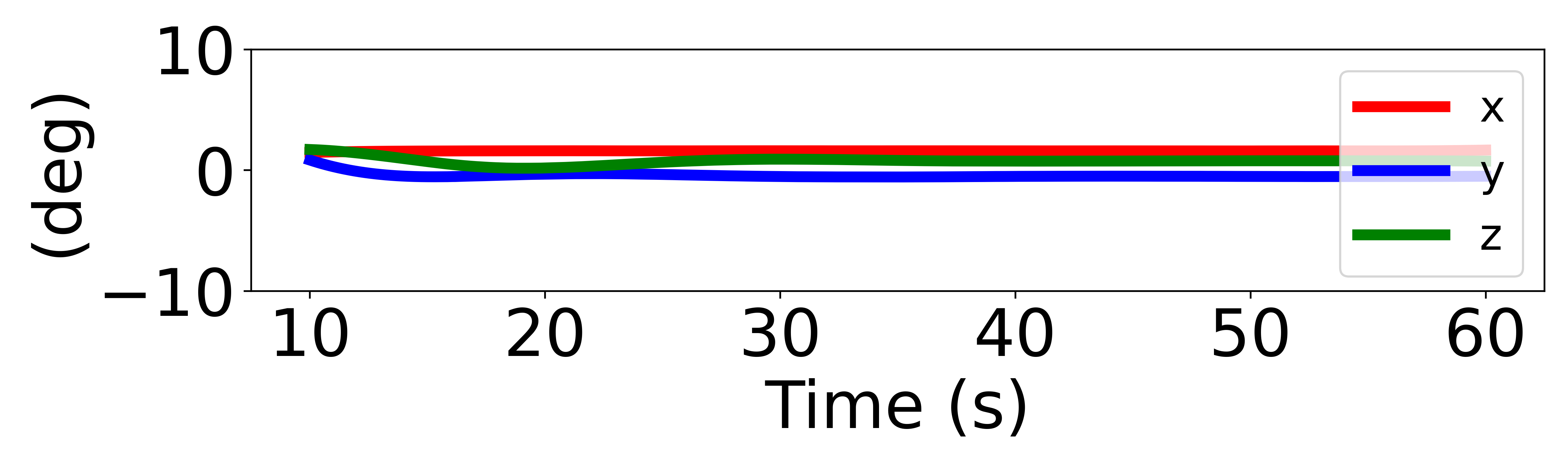}
  \end{subfigure}
  \begin{subfigure}{0.04\textwidth}
    \captionsetup{labelformat=empty}
  \end{subfigure}

  \caption{
    Position and attitude errors under GPS (left) and gyroscope attacks (right). 
    \textbf{Top:} Baseline-RL exhibits significant position and attitude error. 
    \textbf{Middle:} HRP partially mitigates errors but struggles to maintain stable flight. 
    \textbf{Bottom:} \armor maintains significantly lower position and attitude error.
  }
  \label{fig:pos-att-error-comparison}
\end{figure}

Figure~\ref{fig:pos-att-error-comparison} shows the UAV's position error under a GPS spoofing attack. The baseline-RL completely fails under this attack, with a 0\% mission success rate and a 100\% crash rate. 
The trajectory deviation is severe ($>0.9m$), resulting in loss of control.
HRP partially mitigates position errors but struggles to maintain stability, resulting in significant state drift. 
In contrast, \armor maintains accurate position tracking across all axes ($x$, $y$, $z$), keeping the state drift to around 0.1m. 

Figure~\ref{fig:pos-att-error-comparison} also presents the attitude error under a gyroscope attack. 
The baseline-RL exhibits large attitude errors exceeding $\pm$8 degrees, resulting in a 0\% mission success rate, a 100\% crash rate, and a state drift of approximately 0.8 m. 
HRP reduces attitude error, but cannot fully suppress the effects of the attack, resulting in a crash rate of 60\% and a state drift of 18.5 m. 
In contrast, \armor maintains stable attitude control throughout the attack, keeping errors bounded within $\pm$1 degree, resulting in a state drift of less than 3 degrees. 
{\em These results demonstrate \armor's robustness in suppressing different types of physical attacks.}

\begin{table*}[!ht]
\centering
\caption{Performance comparison of HRP, RARL, and \armor under physical attacks against five UAV sensors}
\label{tab:rarl_armor}
\begin{tabular}{l|ccc|ccc|ccc}
\hline
\multirow{2}{*}{\textbf{Target sensor}} & \multicolumn{3}{c|}{\textbf{HRP}}                                                                  & \multicolumn{3}{c|}{\textbf{RARL}}                                                                 & \multicolumn{3}{c}{\textbf{\armor}}                                                               \\ \cline{2-10} 
                                        & \multicolumn{1}{c|}{\textbf{Success}} & \multicolumn{1}{c|}{\textbf{Crash}} & \textbf{State Drift} & \multicolumn{1}{c|}{\textbf{Success}} & \multicolumn{1}{c|}{\textbf{Crash}} & \textbf{State Drift} & \multicolumn{1}{c|}{\textbf{Success}} & \multicolumn{1}{c|}{\textbf{Crash}} & \textbf{State Drift} \\ \hline
GPS                                     & \multicolumn{1}{c|}{40\%}             & \multicolumn{1}{c|}{50\%}           & 6.2 $\pm$ 2.5 m      & \multicolumn{1}{c|}{82\%}             & \multicolumn{1}{c|}{0}              & 0.1 $\pm$ 0.03 m     & \multicolumn{1}{c|}{87\%}             & \multicolumn{1}{c|}{0}              & 0.1 $\pm$ 0.03 m     \\ 
Gyroscope                               & \multicolumn{1}{c|}{32\%}             & \multicolumn{1}{c|}{60\%}           & 18.5 $\pm$ 3.1 deg    & \multicolumn{1}{c|}{78\%}             & \multicolumn{1}{c|}{0}              & 4 $\pm$ 2 deg        & \multicolumn{1}{c|}{83\%}             & \multicolumn{1}{c|}{0}              & 2.3 $\pm$ 1.6 deg    \\ 
Accelerometer                           & \multicolumn{1}{c|}{30\%}             & \multicolumn{1}{c|}{50\%}           & 5.5 $\pm$ 1.7 m/$s^2$    & \multicolumn{1}{c|}{80\%}             & \multicolumn{1}{c|}{0}              & 0.02 $\pm$ 0 m/$s^2$      & \multicolumn{1}{c|}{83\%}             & \multicolumn{1}{c|}{0}              & 0.01 $\pm$ 0 m/$s^2$      \\ 
Magnetometer                            & \multicolumn{1}{c|}{62\%}              & \multicolumn{1}{c|}{15\%}           & 30 $\pm$ 4.1 deg     & \multicolumn{1}{c|}{92\%}             & \multicolumn{1}{c|}{0}              & 8.1 $\pm$ 2.3 deg     & \multicolumn{1}{c|}{94\%}             & \multicolumn{1}{c|}{0}              & 7.7 $\pm$ 2 deg       \\ 
Optical Flow                            & \multicolumn{1}{c|}{46\%}             & \multicolumn{1}{c|}{30\%}           & 7.1 $\pm$ 3.6 m/s      & \multicolumn{1}{c|}{83\%}             & \multicolumn{1}{c|}{0}              & 0.23 $\pm$ 0.05 m/s     & \multicolumn{1}{c|}{90\%}             & \multicolumn{1}{c|}{0}              & 0.1 $\pm$ 0.05 m/s      \\ \hline
\end{tabular}
\end{table*}

Table~\ref{tab:rarl_armor} compares the performance of \armor with two prior techniques:   HRP~\cite{recovery-rl} and RARL~\cite{rarl}, under physical attacks targeting five different sensors. 
Compared to HRP, \armor consistently achieves higher success rates and lower state drift in attacks against all sensor types, while also preventing crashes, significantly outperforming HRP across all the metrics. 
\armor's effectiveness is also higher than that of RARL. 
On average, \armor achieves a higher mission success rate of 88\% compared to RARL’s 83\% and exhibits lower state drift across all attack types. 
{\em Thus, \armor performs better than HRP and RARL under physical attacks.} 

\begin{table}[!ht]
\centering
\footnotesize
\caption{Performance comparison under stealthy attacks.}
\label{tab:stealthy-attacks}
\begin{tabular}{l|c|c|c}
\hline
\textbf{Method} & \textbf{Success (\%)} & \textbf{Crash (\%)} & \textbf{State Drift (m)} \\ \hline
HRP         & 58.5 & 22.0 & 9.1 ± 1.5  \\
RARL        & 63.2 & 18.4 & 7.4 ± 1.2  \\
\armor      & \textbf{80.7} & \textbf{0} & \textbf{3.2 ± 0.9} \\ \hline
\end{tabular}
\end{table}

\subsection{Effectiveness of \armor under Stealthy Attacks}
\label{sec:res-stealthy-attacks}

We evaluate \armor under \emph{stealthy attacks}, where small biases accumulate gradually over time. 
We launch two types of stealthy attacks~\cite{stealthy-attacks, specguard} targeting the GPS of the UAV: 
(1) \emph{ramp profiles}, where the injected bias grows continuously at a fixed rate 
\((f(t) = f_0 + r t)\), and 
(2) \emph{stair profiles}, where the bias increases in discrete increments at fixed intervals 
\((f(t_k) = f(t_{k-1}) + \delta)\). 
The ramp rates and bias increments chosen all fall within the bounds of Table~\ref{tab:attack-params}. 

Our results in Table~\ref{tab:stealthy-attacks} show that \armor achieves higher success,  and lower state drift, compared to HRP and RARL, which indicates its resilience to stealthy attacks. It also incurs no crashes. 
This resilience arises because \armor’s attack-aware latent representation, combined with its history-based encoder, captures gradual bias trends, which prevent small errors from compounding into large deviations.

\subsection{Zero-Shot Performance}
\label{sec:res-zero-shot}

We evaluate \armor's effectiveness against attacks {\em not encountered during training}. Table~\ref{tab:zeroshot-gps-gyro} and Table~\ref{tab:zeroshot-gyro-gps} compare the zero-shot performance of \armor and RARL, the current state-of-the-art approach for adversarially training robust policies. 
Specifically, we evaluate control policies trained exclusively on a single attack type (either GPS or gyroscope) and tested on \textit{unseen attacks} targeting a different sensor.

\begin{table}[!ht]
\centering
\footnotesize
\caption{Zero-shot performance of RARL and \armor when trained on GPS manipulations only and tested on unseen attacks (Gyroscope and Gyroscope+Accelerometer).}
\label{tab:zeroshot-gps-gyro}
\begin{tabular}{ll|c|c}
\hline
\multicolumn{2}{c|}{\textbf{Metrics}}                      & \multicolumn{1}{c|}{\textbf{Gyroscope}} & \multicolumn{1}{c}{\textbf{Gyro+Accelerometer}} \\ \hline
\multicolumn{1}{c|}{\multirow{3}{*}{RARL}}   & Success     & 0\%                                       & 0\%                                           \\  
\multicolumn{1}{c|}{}                        & Crash       & 60\%                                    & 75\%                                        \\  
\multicolumn{1}{c|}{}                        & State Drift & 15$\pm$ 5.1 deg                         & 12.3$\pm$2.6 deg, 8$\pm$2.1 m/$s^2$                              \\ \hline
\multicolumn{1}{c|}{\multirow{3}{*}{\armor}} & Success     & 60\%                                    & 50\%                                        \\  
\multicolumn{1}{c|}{}                        & Crash       & 0\%                                       & 10\%                                           \\  
\multicolumn{1}{c|}{}                        & State Drift & 3.5 $\pm$ 1.8 deg                       & 2.8$\pm$1.6 deg, 1.1$\pm$0.4 m/$s^2$                            \\ \hline
\end{tabular}
\end{table}

\vspace{-5mm}

\begin{table}[!ht]
\centering
\footnotesize
\caption{Zero-shot performance of RARL and \armor when trained on Gyroscope manipulations only and tested on unseen attacks (GPS and GPS+Accelerometer).}
\label{tab:zeroshot-gyro-gps}
\begin{tabular}{ll|c|c}
\hline
\multicolumn{2}{c|}{\textbf{Metrics}}                      & \multicolumn{1}{l|}{\textbf{GPS}} & \multicolumn{1}{l}{\textbf{GPS+Accelerometer}} \\ \hline
\multicolumn{1}{c|}{\multirow{3}{*}{RARL}}   & Success     & 5\%                                   & 5\%                                         \\  
\multicolumn{1}{c|}{}                        & Crash       & 70\%                                  & 80\%                                        \\  
\multicolumn{1}{c|}{}                        & State Drift & 6.5 $\pm$ 2.2 m                        & 11.5 $\pm$ 2.1 m, 10.2 $\pm$ 2.7 m/$s^2$         \\ \hline
\multicolumn{1}{c|}{\multirow{3}{*}{\armor}} & Success     & 70\%                                   & 55\%                                        \\  
\multicolumn{1}{c|}{}                        & Crash       & 5\%                                    & 8\%                                         \\  
\multicolumn{1}{c|}{}                        & State Drift & 0.6 $\pm$ 0.2 m                        & 0.8 $\pm$ 0.3 m, 1.5 $\pm$ 0.7 m/$s^2$          \\ \hline
\end{tabular}
\end{table}
There are two cases we considered. 

\textbf{Case 1:}  Policies trained on GPS manipulations, which introduce drift biases in position estimates, are evaluated on unseen gyroscope attacks that induce high-frequency oscillatory biases, and on multi-sensor (gyroscope+accelerometer) attacks. We find that 
RARL achieves 0\% success and a 60\% crash rate under gyroscope attacks. 
For multi-sensor attacks, RARL's performance further deteriorates, with a 75\% crash rate, and significant state drift. 
In contrast, \armor achieves a 60\% success rate under gyroscope attacks and 50\% under multi-sensor attacks, while maintaining crash rates below 10\% and reducing state drift by $4\times$ compared to RARL.

\textbf{Case 2:} Policies trained on gyroscope manipulations, which introduce high-frequency oscillatory biases, are evaluated on unseen GPS attacks that induce slow drift biases, as well as multi-sensor (GPS + accelerometer) attacks. We find that 
RARL generalizes poorly, achieving only 5\% success and exhibiting high crash rates (70--80\%) and significant state drift. 
In contrast, \armor shows strong zero-shot generalization, achieving a success rate of 70\% under GPS attacks and 55\% under multi-sensor attacks, while maintaining low crash rates, and reducing state drift by over 10$\times$.

{\em These results highlight \armor's ability to generalize to unseen attacks targeting both single and multiple sensors}.

%% file: sections/6-discussions.tex

\armor offers two main advantages over adversarial training: (1) training efficiency, and (2) zero-shot effectiveness. 

Adversarial training methods~\cite{rarl, adv-agent-training, adv-train-deeprl} involve iterative policy updates by co-training an antagonist policy to generate adversarial perturbations. This results in high computational costs and long training times. 
In contrast, \armor’s two-stage training framework removes the need for an explicit antagonist. 
Instead, it leverages attack-aware latent state representations during training, and transfers the knowledge to a student encoder for deployment.
As shown in Figure~\ref{fig:training-comparison}(b), \armor achieves comparable effectiveness to RARL while requiring significantly fewer training timesteps. 

Furthermore, \armor demonstrates strong zero-shot generalization capabilities, enabling the control policy to handle unseen attack types, including both single-sensor and multi-sensor attacks. 
\armor significantly outperforms RARL in zero-shot evaluations, achieving higher success rates, significantly lower crash rates, and reduced state drift. 

{\em \textbf{Limitations.}}
While \armor demonstrates promising zero-shot robustness to single unseen attack types, generalization to multi-sensor attacks is limited due to the compounding effects of the perturbations. This is an avenue for future work.

Our experiments focus on evaluation across diverse attack types targeting different sensors within a single UAV model and simulator; extending this to other robotic platforms and sim-to-real transfer is an important direction for future work.

%% file: sections/7-conclusions.tex
We introduced \armor, a two-stage learning framework for attack-resilient UAV control. 
\armor leverages attack-aware privileged information during training to learn robust latent state representations, and uses transfer learning to adapt these representations for online deployment. 
This approach eliminates the need for iterative adversarial training, resulting in a more efficient and scalable framework.
Our results demonstrate that \armor maintains safe and stable flight, outperforming existing techniques. 
Furthermore, \armor exhibits promising zero-shot generalization, enabling resilience against previously unseen attacks.
Future work will explore extending \armor to a broader class of robotic systems. 
We will also integrate theoretical safety guarantees, and constraint satisfaction under adversarial conditions.